\documentclass[lettersize,journal]{IEEEtran}
\usepackage{amsmath,amsfonts}
\usepackage{algorithmic}
\usepackage{algorithm}
\usepackage{array}
\usepackage[caption=false,font=normalsize,labelfont=sf,textfont=sf]{subfig}
\usepackage{textcomp}
\usepackage{stfloats}
\usepackage{url}
\usepackage{verbatim}
\usepackage{graphicx}
\usepackage{epstopdf}
\usepackage{cite}
\usepackage{bm}
\usepackage{caption}
\usepackage{booktabs}
\usepackage{diagbox}
\usepackage{subcaption}

\begin{document}

\title{On-site scale factor linearity calibration of MEMS triaxial gyroscopes}


\author{
	\IEEEauthorblockN{Yaqi Li$^{a}$, Li Wang$^{a}$, Zhitao Wang$^{a}$, Xiangqing Li$^{a}$, Jiaojiao Li$^{b}$, Steven Weidong Su$^{a*}$}
	
	\IEEEauthorblockA{$^{a}$ College of Artificial Intelligence and Big Data for Medical Sciences, Shandong First Medical University Shandong Academy of Medical Sciences, 6699 Qingdao Road, Jinan, 250024, Shandong, China
	}
	W
	\IEEEauthorblockA{	$^{b}$ Faculty of Engineering and IT, University of Technology Sydney, Australia
	}
	
	\thanks{* Corresponding Author}
}
\markboth{Journal of \LaTeX\ Class Files,~Vol.~14, No.~8, August~2021}%
{Shell \MakeLowercase{\textit{et al.}}: A Sample Article Using IEEEtran.cls for IEEE Journals}


\maketitle

\begin{abstract}
The calibration of MEMS triaxial gyroscopes is crucial for achieving precise attitude estimation for various wearable health monitoring applications. However, gyroscope calibration poses greater challenges compared to accelerometers and magnetometers. This paper introduces an efficient method for calibrating MEMS triaxial gyroscopes via only a servo motor, making it well-suited for field environments. The core strategy of the method involves utilizing the fact that the dot product of the measured gravity and the rotational speed in a fixed frame remains constant. To eliminate the influence of rotating centrifugal force on the accelerometer, the accelerometer data is measured while stationary. The proposed calibration experiment scheme, which allows gyroscopic measurements when operating each axis at a specific rotation speed, making it easier to evaluate the linearity across a related speed range constituted by a series of rotation speeds. Moreover, solely the classical least squares algorithm proves adequate for estimating the scale factor, notably streamlining the analysis of the calibration process. Extensive numerical simulations were conducted to analyze the proposed method's performance in calibrating a triaxial gyroscope model. Experimental validation was also carried out using a commercially available MEMS inertial measurement unit (LSM9DS1 from Arduino nano 33 BLE SENSE) and a servo motor capable of controlling precise speed. The experimental results effectively demonstrate the efficacy of the proposed calibration approach.
\end{abstract}


\section{Introduction}
\IEEEPARstart{T}{he} Micro-Electro-Mechanical Systems (MEMS) find wide-ranging applications in daily life owing to their affordability, compact size, and low power consumption, evident in devices like health monitoring equipment\cite{s17112573}\cite{7892837}\cite{wang2021efficient}, smartphones\cite{huang2020combination}, and personal consumer electronic devices\cite{han2016extended}\cite{passaro2017gyroscope}. However, compared to expensive high-precision gyroscopes (such as fiber optic gyroscopes), MEMS face challenges regarding accuracy stability that cannot be overlooked. Numerous factors contribute to MEMS accuracy, with sensitivity to environmental conditions and structural design constraints due to size limitations being paramount. For instance, fluctuations in ambient temperature can significantly alter the scale factor and bias of MEMS gyroscopes\cite{s18093004}\cite{7061397}. One effective mitigation strategy involves performing rapid calibration prior to usage. Nonetheless, calibrating MEMS gyroscopes proves more intricate compared to accelerometers and magnetic sensors. Gyroscope calibration primarily aims at mitigating system errors, predominantly stemming from scale factor and bias discrepancies, with bias being relatively easier to rectify. Thus, gyroscope calibration predominantly focuses on scale factor calibration, particularly emphasizing the assessment of linearity stability within the operational range, posing a formidable yet inevitable challenge.

The calibration issue of gyroscope has garnered widespread attention from both academia and industry\cite{s18041123}. The most original method of triaxial gyroscope calibration required high installation accuracy and expensive equipment\cite{FGA} \cite{hall2000case}. The requirement of equipment and complex process make it not suitable for the field environment. In \cite{wu2017gyroscope}, The magnetometer data was utilized to calibrate the gyroscope, requiring neither high precision nor expensive equipment. However, the magnetic field is so weak that it is easily influenced by external alternating magnetic fields. The other paper \cite{9739788} , use the temporal convolutional network to calibrate the MEMS gyroscope, which can calibrate the gyroscope well but it is not suitable for the edge computing devices due to its large and complex neural networks and calculation of non-linearity. And in \cite{9499084},  the sum of the squares of the three axis is equal to 1 to correct the scale factor of the gyroscope, which includes the nonlinear least squares. The nonlinear least squares may cause results not to converge. In \cite{6413274}, the camera was employed to calibrate the gyroscope which did not require the high precise equipment but require high mounting accuracy. The method proposed by \cite{5412920} aims to obviate the requirement for a turntable in the gyroscope's calibration process. They combined accelerometer with a bicycle wheel to calibrate the gyroscope. This method can calibrate the accelerometer and gyroscope in filed but only two axes of the gyroscope can be calibrated. In \cite{glueck2015automatic}, a novel multi-model error state Kalman filter is presented for estimating the gyroscope's offset. However, the method involves a substantial number of nonlinear operations, rendering the calibration process more intricate.

In this paper, we introduce a new triaxial gyroscope calibration method utilizing linear least squares to prevent estimation divergence. The underlying principle of this method relies on the constancy of the dot product between a fixed rotation vector and the local gravity vector. The calibration process requires only a low-cost servo motor, with a tolerance within 1\% when equipped with an encoder. Additionally, this method necessitates only a one-time installation on the servo motor. These advantages render this calibration approach highly efficient for on-site triaxial gyroscope calibration.

The conventional gyroscope calibration methods consist of the 9-parameter approach, which incorporates installation misalignment error, and the 6-parameter approach, which excludes it \cite{6561506}. In this study, we adopt the 6-parameter method for gyroscope calibration, and our proposed approach accurately estimates scale factors even in the presence of installation misalignment.

To address the influence of rotating centrifugal force on the accelerometer, accelerometer data is acquired while stationary. Subsequently, to mitigate the impact of accelerometer errors on calibration outcomes, we employ the calibration method proposed by L. Ye \cite{7946167}. This method effectively calibrates the accelerometer and reduces the influence of accelerometer errors on calibration results.

As previously discussed, acquiring gyroscope data at a single rotational speed for scale factor calibration eliminates the need for reinstalling the gyroscope; it only needs to be installed once \cite{bistrov2012performance}. Furthermore, since scale factor calibration occurs at one rotational speed with a single installation, this characteristic can be leveraged to deduce the gyroscope's linearity within the specified range by selecting a series of relevant speeds. Moreover, the parameter estimation involves only classical linear least squares operation, facilitating the determination of a triaxial scale factor. This simplifies the investigation of the linearity of the scale factor for each axis, making it more effective. 

We summarize the contributions as follows: Firstly, this paper introduces a highly efficient method for calibrating the triaxial gyroscope, capable of determining its linearity within the specified interval. Secondly, the proposed method enables simultaneous calibration of all three axes with only one installation, effectively mitigating the impact of installation errors resulting from multiple installations on the final calibration results. Thirdly, the calibration method involves only the least squares method, which is computationally simple and facilitates high-frequency calibration. Finally, this calibration method does not necessitate an expensive high-precision turntable; instead, it only requires a servo motor or an industrial robot arm.


\section{calibration methodology}

In this section, we will introduce the proposed calibration procedure with an emphasis on the underlying mathematical principles. The performance of a gyroscope is influenced by various factors, with a particular focus on analyzing the scale factor. Introducing $\mathbf{G^{r}}$ as the calibrated angular velocity, $\mathbf{G^{m}}$ as the uncalibrated angular velocity, and $\mathbf{b}$ as bias, we can express the calibration model as follows:
\begin{equation} \label{eq1}
	\mathbf{G^{r}=K*G^{m}+b}
\end{equation}
Here, the calibration methods encompass two models: the 9-parameter and 6-parameter model respectively. The difference between the two models lies in $\mathbf{K}$, which can be described as follows:

For the 9-parameter model:
$$\begin{bmatrix}
k_{xx} & k_{xy} & k_{xz} \\
k_{yx} & k_{yy} & k_{yz} \\
k_{zx} & k_{zy} & k_{zz} \\
\end{bmatrix}.$$

For the 6-parameter model: 
$$\begin{bmatrix}
  k_{xx} & 0 & 0 \\
  0 & k_{yy} & 0 \\
  0 & 0 & k_{zz} \\
  \end{bmatrix}.$$

However, in this study, we primarily focused on the calibration of scale factors. We assume that the installation error compared to the scale factor is negligible, i.e., $k_{xy}=k_{xz}=k_{yx}=k_{yz}=k_{zx}=k_{zy}=0$. Additionally, even if installation errors exist, the proposed method can still achieve the desired calibration accuracy for the scale factors.

\begin{figure}
	\centering
	\captionsetup{justification=centering}
	\includegraphics[width=0.4\textwidth,height=0.4\textwidth]{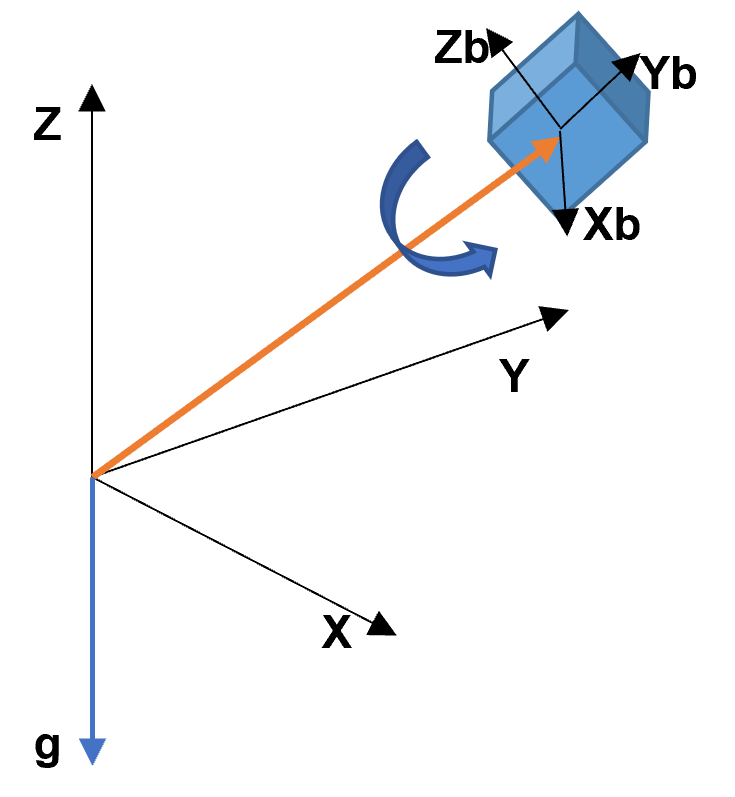}
	\caption{The alignment of the IMU, local gravity, and rotational axis during the calibration process.}
	\label{fig_ort}
\end{figure}

Now, let's introduce the calibration procedure. Firstly, we install the MEMS IMU on a servo motor as shown in Figure \ref{fig_ort}.

Before powering the motor, we measure the bias and remove them from the triaxial gyroscopes under steady state. Then, we rotate the motor by hand to at least three static positions and record the readings of the triaxial accelerometers.

Next, we power on the motor and let it rotate around the same axis at a relatively constant speed, while recording the readings of the triaxial gyroscopes. It is important to note that during these rotations, the angle between the rotation axis and gravity remains constant.

Based on the above procedure, considering the dot product between the gyroscope vector and the accelerometer vector, the constant $L$ can be expressed as follows:

\begin{equation} \label{eq2}
  L = \mathbf{a}\mathbf{g^{r}} = 
  \begin{bmatrix}
    A_{x} & A_{y} & A_{z}
  \end{bmatrix}
  \begin{bmatrix}
    G_{x}^{r} \\ G_{y}^{r} \\ G_{z}^{r}
  \end{bmatrix}
  = A_{x}G_{x}^{r} + A_{y}G^{r}_{y} + A_{z}G_{z}^{r}
\end{equation}
To ensure the accuracy of the calibration results, the data $ \mathbf{a^{m}} = \begin{bmatrix}
  A^{m}_{x} & A^{m}_{y} & A^{m}_{z}
\end{bmatrix}^T $ from the measured accelerometer will be calibrated using the method proposed in our previous work \cite{7946167}. The calibration results are identified as $\mathbf{a^{c}} = \begin{bmatrix}
  A^{c}_{x} & A^{c}_{y} & A^{c}_{z}
\end{bmatrix}^T $. Based on (\ref{eq1}), the constant $L$ can is shown below:
\begin{equation} \label{eq3}
  L = A_{x}^{c}(K_{x}G^{m}_{x}+b_{x})+A_{y}^{c}(K_{y}G^{m}_{y}+b_{y})+A_{z}^{c}(K_{z}G^{m}_{z}+b_{z})
\end{equation} 
In (\ref{eq3}), the bias 
$\mathbf{b} = \begin{bmatrix}b_{x}&b_{y}&b_{z}\end{bmatrix}^T$ can be easily eliminated. It can be rewritten in matrix form, when there were $n$ sets of data, it can be expressed as follows:
 
\begin{equation} \label{eq4}
  \begin{bmatrix}
    A_{x1}^{c}G_{x}^{m}&A_{y1}^{c}G_{y}^{m}& A_{z1}^{c}G_{z}^{m} \\
    \cdots & \cdots & \cdots\\
    \vdots & \vdots & \vdots \\
    A_{xn}^{c}G_{x}^{m}&A_{yn}^{c}G_{y}^{m}&A_{zn}^{c}G_{z}^{m}\\
  \end{bmatrix}
  \begin{bmatrix} K_{x}\\K_{y}\\ K_{z} \end{bmatrix}  = \begin{bmatrix}
      L \\\vdots\\L
  \end{bmatrix}
\end{equation} 
Equation (\ref{eq4}) can be simplified as below:   
\begin{equation} \label{eq5}
  \bm{l = X\beta + \epsilon}
\end{equation}
where, $$ \mathbf{X} = 
\begin{bmatrix}
  A_{x1}^{c}G_{x}^{m}&A_{y1}^{c}G_{y}^{m}& A_{z1}^{c}G_{z}^{m} \\
  \cdots & \cdots & \cdots\\
  \vdots & \vdots & \vdots \\
  A_{xn}^{c}G_{x}^{m}&A_{yn}^{c}G_{y}^{m}&A_{zn}^{c}G_{z}^{m}\\
\end{bmatrix}$$
 and $\bm{\beta} = \begin{bmatrix} K_{x} & K_{y} & K_{z} \end{bmatrix}^{T}$, and $\epsilon$ represents the measurement noise. Then, the parameters to be calibrated can be expressed as follows:
 \begin{equation} \label{eq6}
 	\bm{\beta = (X^{T}X)^{-1}X^{T}l}.
 \end{equation}

In this framework, we assert that the square of a gyroscope's rotational speed is equivalent to the sum of the squares of its readings across three axes. This foundational principle allows us to express the relationship as follows:

\begin{equation} \label{eq7}
	(G_{x}^{m}\hat{K}_{x}\alpha)^{2} +(G_{y}^{m}\hat{K}_{y}\alpha)^{2}+(G_{z}^{m}\hat{K}_{z}\alpha)^{2} = n^{2}
\end{equation}

In this equation, \(n\) represents the servo motor's rotational speed, which can be well controlled within an experiment. The servo motor's speed error can often be maintained below 1\%, facilitating accurate calibration. Upon calculating \(\alpha\), the calibrated scale factor \(K\) is derived as \(\alpha\) times the transpose of the initial scale factor estimates vector \(\begin{bmatrix} \hat{K}_{x} & \hat{K}_{y} & \hat{K}_{z} \end{bmatrix}^{T}\). This calibrated scale factor \(K\) represents a refined adjustment to the gyroscope's scale factors, enhancing measurement accuracy by aligning them with the known servo motor speed.

The aforementioned approach facilitates rapid and straightforward calibration of a specific rotating speed point of the gyroscope, yielding its scale factor. Expanding upon this method enables calibration across multiple speed points in the gyroscope's operational range. By densely computing the scale factors for various speed points, the gyroscope's linearity can be determined.

\section{Simulation Study}

In this section, simulation results will be presented to test and verify the effectiveness of the proposed calibration method.

In the simulation, coordinate transformations will be employed to emulate the installation posture of the IMU and its associated rotating axis. Quaternions will be used to simulate the axis of rotation.

\subsection{Simulation at a specific speed}
Initially, simulations are conducted at a specified rotational speed with certain parameter settings. The configuration is as follows:

\begin{itemize}
	\item The scale factor is chosen according to a uniform distribution, $U(0.9, 1.1)$. This reflects the typical variability of MEMS gyroscope scale factors, which usually falls within $\pm 10\%$.
	\item The bias parameter follows a uniform distribution, $U(-3^\circ/s, 3^\circ/s)$, corresponding to the expected range of MEMS gyroscope biases of $[-3 ^\circ/s, 3^\circ/s]$.
	\item Measurement noise is modeled by a Gaussian distribution, $\mathcal{N}(0, 0.1^{2})$, to simulate realistic sensor noise conditions.
	\item Installation alignment issues are disregarded, assuming that the experiment's setup does not necessitate precise mounting alignment.
\end{itemize}

Based on these parameter settings, simulation data are generated, and the calibration results will subsequently be discussed. 

In this simulation, the selected scale factor is $
\begin{bmatrix}
    K^{s}_{x} & K^{s}_{y} & K^{s}_{z}
\end{bmatrix}^T =
\begin{bmatrix}
    1.033 & 0.811 & 1.151
\end{bmatrix}^T$
and the axis rotation vector is 
$\begin{bmatrix}
    \omega_x & \omega_y & \omega_z
\end{bmatrix}^T = 
\begin{bmatrix}
    -1 & 1 & -1
\end{bmatrix}^T$.

\begin{figure}
    \centering
    \captionsetup{justification=centering}
    \includegraphics[width=0.5\textwidth,height=0.25\textwidth]{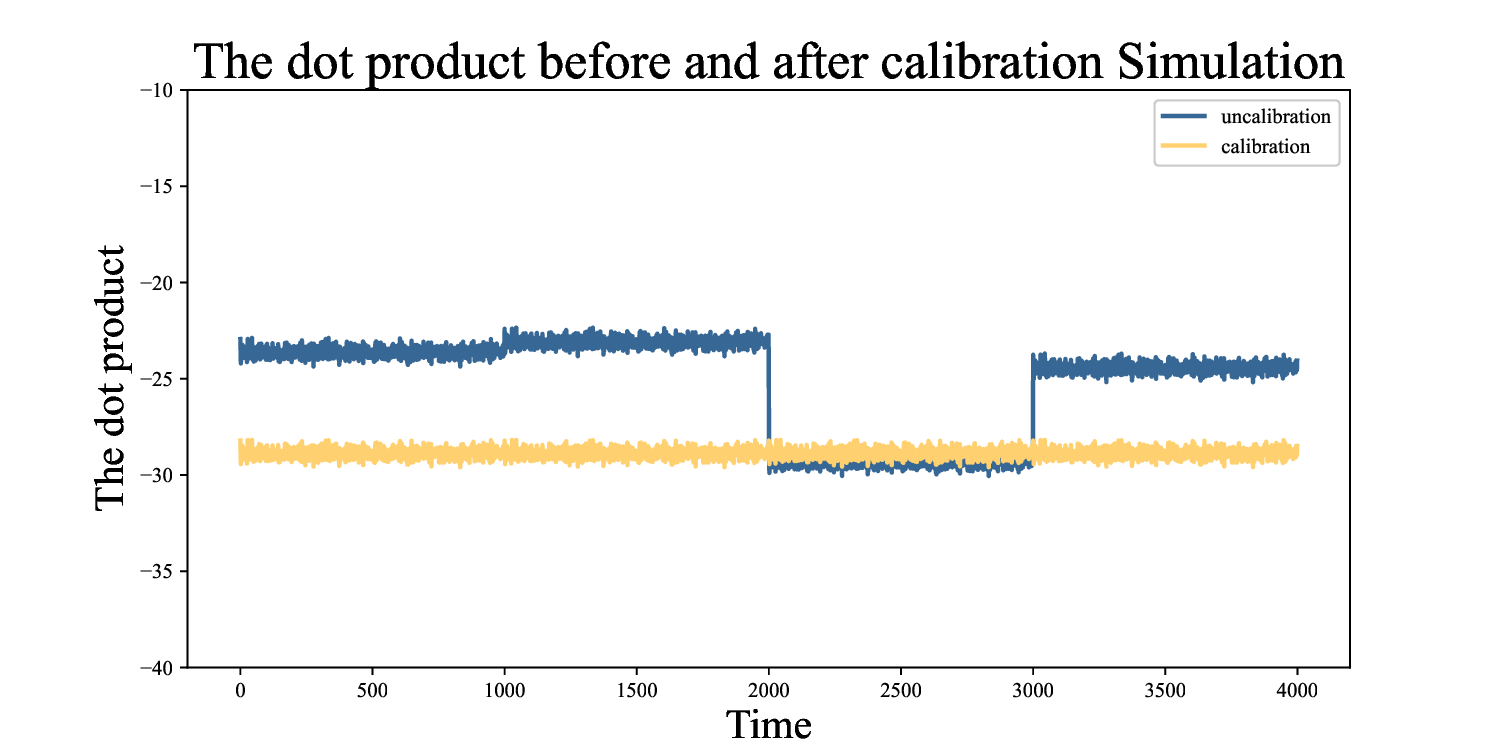}
    \caption{The dot product between the gyroscope and the accelerometer before and after calibration}
    \label{fig:process}
\end{figure}

After setting the rotational speed to $50^{\circ}/s$, the system underwent the calibration process. Following the calibration procedure, the outcomes are illustrated in Figure \ref{fig:process}. Initially, there is a noticeable discrepancy between the dot product of the gyroscope and accelerometer vectors. Post-calibration, this inconsistency is resolved, leading to a uniform alignment.

To verify the repeatability of this methodology, 500 parameter sets were created for the statistic analysis of gyroscope calibration. A boxplot is used to present the results of the calibration. Additionally, scale factors, bias, noise, and other variables adhere to the established guidelines, and the rotational speed is set at $50^{\circ}/s$.

\begin{figure}
    \centering
    \captionsetup{justification=centering}
    \includegraphics[width=0.48\textwidth,height=0.36\textwidth]{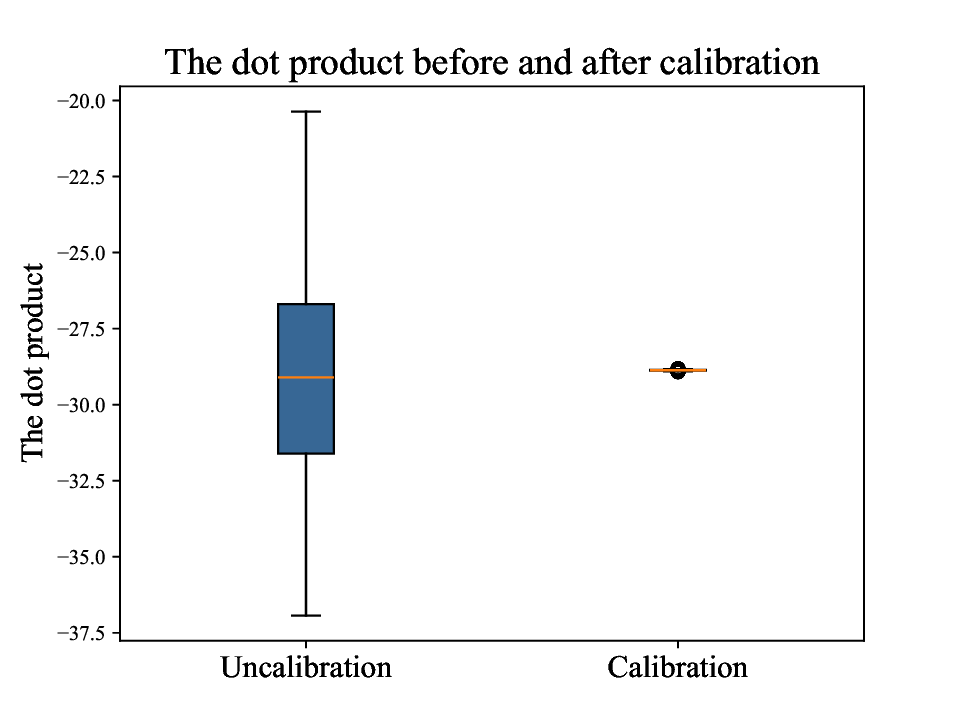}
    \caption{The statistic results of Calibration and Uncalibration}
    \label{fig:statistic_result}
\end{figure}
As illustrated in Figure \ref{fig:statistic_result}, it is evident that before calibration, the distribution of dot products is scattered; however, after calibration, the distribution of the product becomes concentrated. This observation demonstrates the effectiveness of the calibration process in reducing the variance and improving the consistency of dot product calculations.

\subsection{Simulation for assessing linearity}
Assessing the linearity of a gyroscope across specific rotational speeds efficiently and effectively remains a significant challenge. This subsection is dedicated to evaluating the gyroscope's linearity through simulation. 
This involves generating random scale factors for the gyroscope at various speed points to mimic the real-life phenomenon of random drift, then employing a proposed calibration method to assess linearity across a defined interval.

\subsection*{Steps for the Simulation}

\begin{figure}
	\centering
	\captionsetup{justification=centering}
	\includegraphics[width=0.48\textwidth,height=0.36\textwidth]{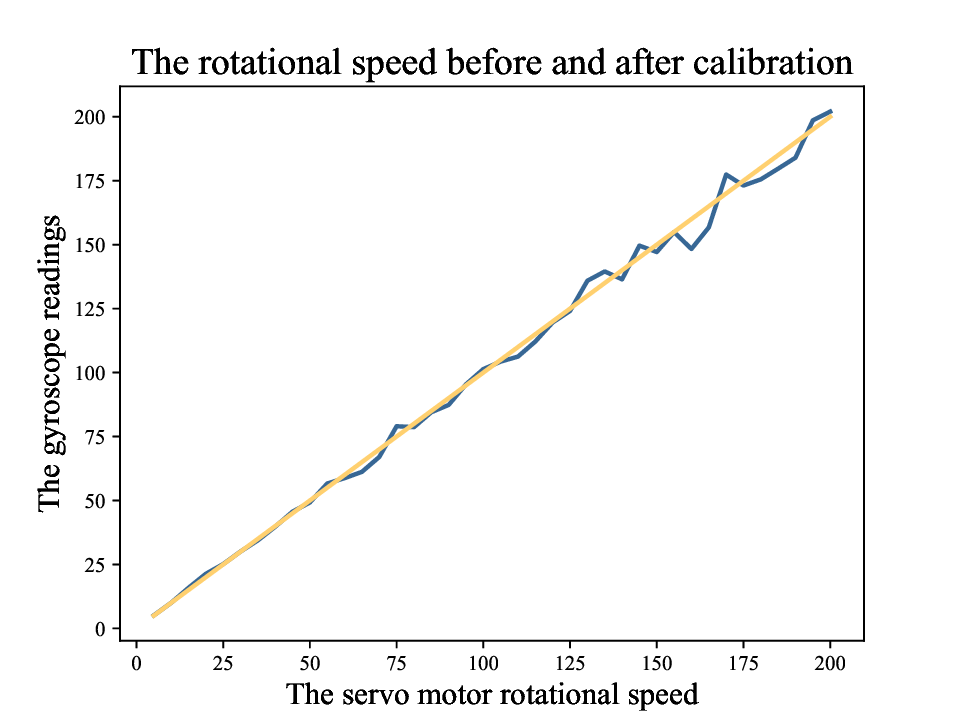}
	\caption{The speed before and after calibration}
	\label{fig:linearity_speed}
\end{figure}

\begin{enumerate}
	\item \textbf{Generate Random Scale Factors:} Simulate the gyroscope's performance by generating random scale factors at each speed point within the range of $0^{\circ}/s$ to $200^{\circ}/s$, with increments of $5^{\circ}/s$.
	\item \textbf{Apply Calibration Method:} Use the proposed calibration method to correct for any non-linearity observed due to the random drift.
	\item \textbf{Evaluate Linearity:} Through the corrected scale factors, assess the gyroscope's linearity over the specified speed range.
	\item \textbf{Plotting:} Visualize the calibration results with a plot showing the gyroscope's performance (both pre and post-calibration) against the actual rotational speeds of the servo motor.
\end{enumerate}
The calibration outcomes, as depicted in Figure \ref{fig:linearity_speed}, demonstrate that the gyroscope's readings, post-calibration, closely mirror the actual rotational speeds produced by the servo motor. This finding demonstrates the efficacy of the proposed calibration method in aligning the gyroscope's output with real-world rotational speeds.

It is well-documented that servo motors produce vibrations as they rotate, as noted in \cite{svetlitsky2012engineering}. These vibrations, along with noise emissions from gyroscopes, play a substantial role in affecting the accuracy of calibration results.

This study intends to investigate the influence of noise on calibration outcomes through a comprehensive series of simulations. These simulations will assess how different levels of noise impact the calibration accuracy at various rotational speeds. To ensure the reliability of the findings, each simulation scenario will be replicated a thousand times. The results will be depicted using 2D box plots and 3D bar plots, facilitating a clear visual representation of how noise levels correlate with calibration precision.

The simulation is configured with the following parameters: the rational speed spans from $5^{\circ}/s$ to $200^{\circ}/s$, while noise arising from servo motor vibration and MEMS is represented by Gaussian noise ranging from $\mathcal{N}(0, 0^{2})$ to $\mathcal{N}(0, 200^{2})$.

To illustrate the influence of various noise magnitudes on calibration outcomes, we assume constant gyroscope drift across all speed points in the simulation. The assumed scale factor is $\begin{bmatrix}
    k_x & k_y & k_z
\end{bmatrix}^T$ =$\begin{bmatrix}
    1.05 & 0.95 & 1.1
\end{bmatrix}^T$.

\begin{figure}
    \centering
    \captionsetup{justification=centering}
    \includegraphics[width=0.45\textwidth,height=1\textwidth]{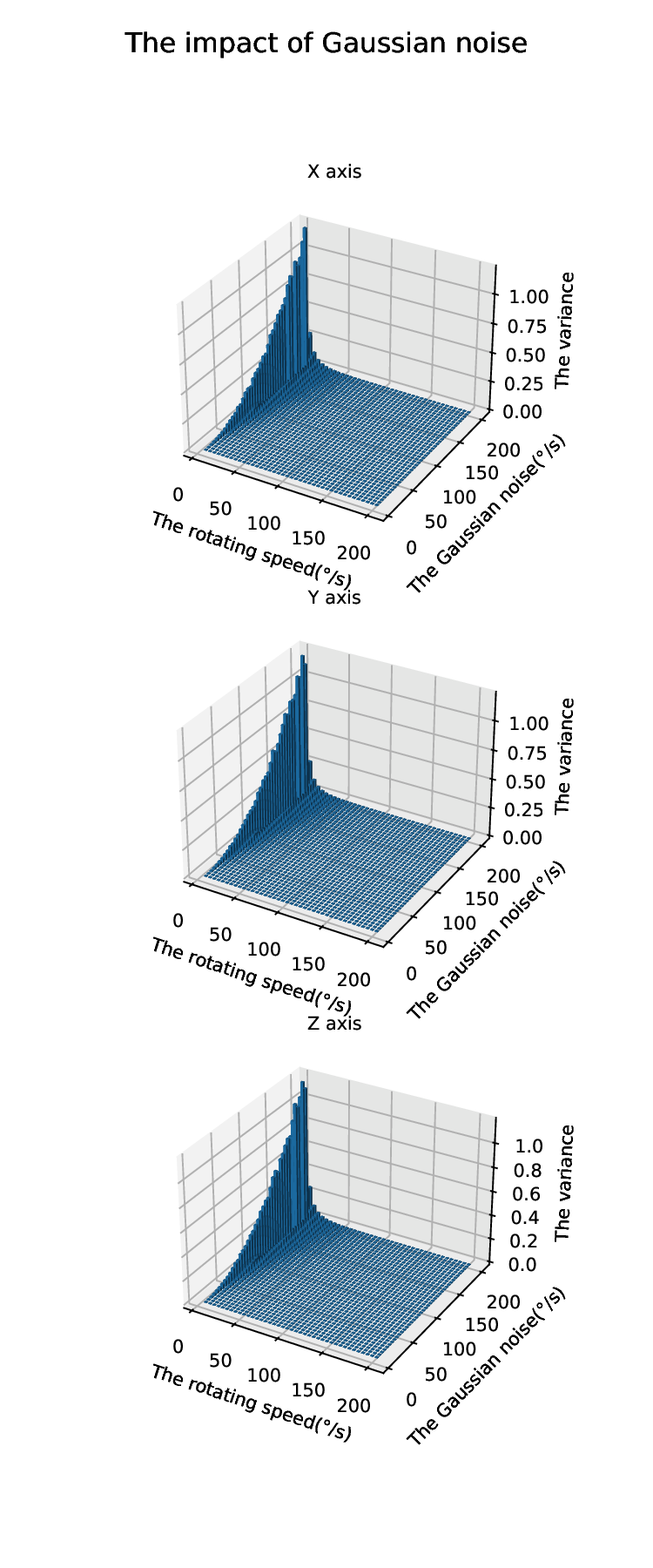}
    \caption{The impact of different sizes of Gaussian noise on calibration results at different speeds }
    \label{fig:3d}
\end{figure}

\begin{figure*}
    \centering
    \captionsetup{justification=centering}
    \includegraphics[width=1\textwidth,height=0.9\textwidth]{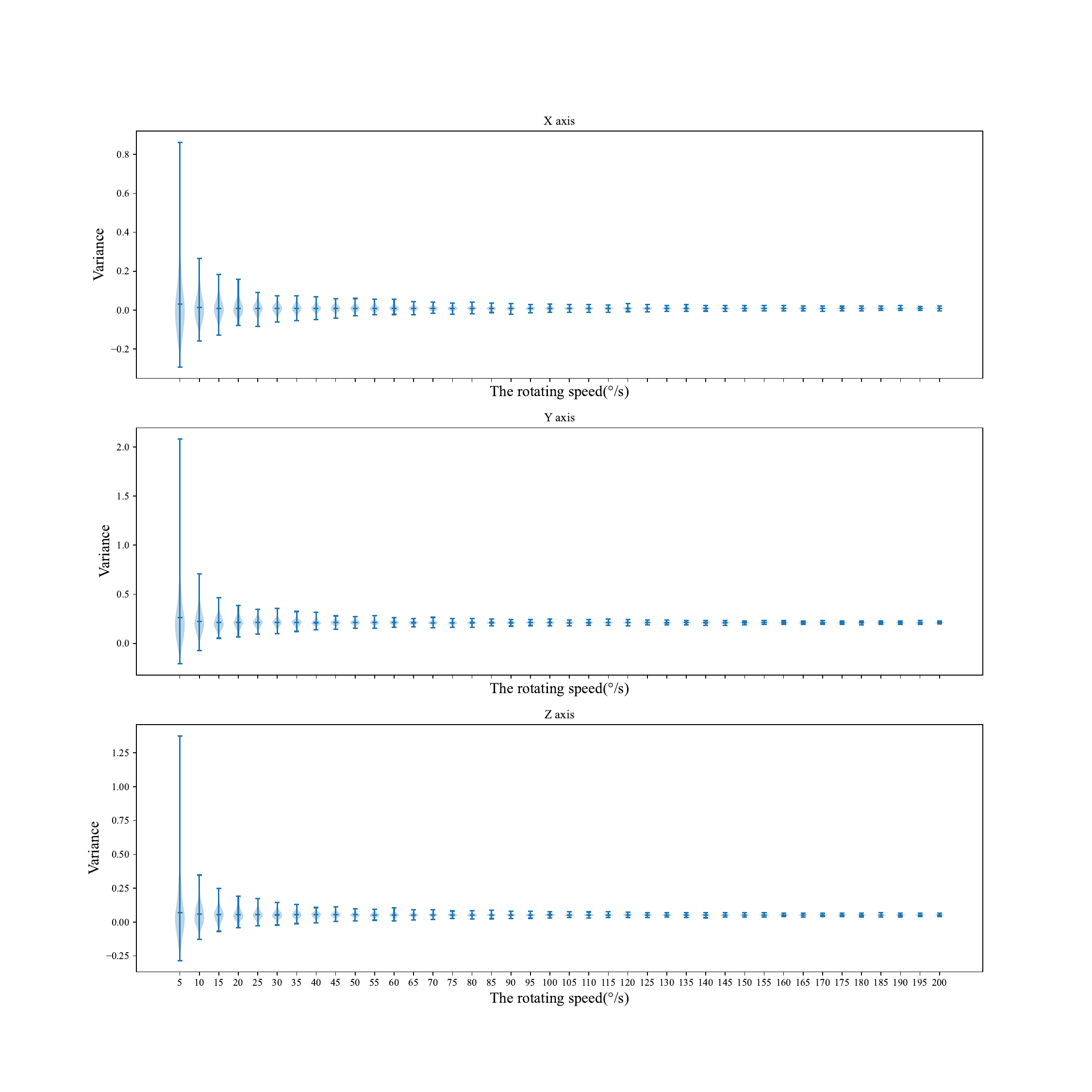}
    \caption{The impact of Gaussian noise on calibration results at different rotating speed point}
    \label{fig:var}
\end{figure*}

The calibration outcomes are presented in  Figure \ref{fig:3d} and \ref{fig:var}. Analyzing Figure \ref{fig:3d}, it can be deduced that at a consistent speed, there's a noticeable increase in the variance of calibration results corresponding to elevated levels of noise. Conversely, in scenarios where noise levels are held constant, higher rotational speeds exhibit a mitigated influence on calibration outcomes, as depicted in Figure \ref{fig:var}. 

To effectively show the impact of noise on calibration outcomes throughout the calibration procedure, we conducted a simulation across the rotational speed ranging from $5^{\circ}/s$ to $200^{\circ}/s$, with Gaussian noise set at $\mathcal{N} (0,30^{2})$. Employing a violin plot, our analysis reveals a distinct trend: as rotational speed increases, the influence of noise on calibration outcomes gradually diminishes. Particularly at lower rotating speeds, the scale factor derived from the initial calibration may be susceptible to higher levels of noise. Consequently, it is advisable to repeat the calibration process at low speeds multiple times and average the results to ensure heightened reliability.

As rotational speed transitions to medium and high levels, the impact of noise on calibration outcomes gradually diminishes, facilitating more accurate results with a single calibration iteration. 

In conclusion, the proposed method effectively calibrated the scale factor for each axis of the gyroscope in the simulation, offering an intuitive depiction of the gyroscope's linearity within a defined interval. This calibration process enhances the accuracy of gyroscope measurements, ensuring reliable performance across various operating conditions.


\section{Experiment}
\begin{figure}[H]
    \centering
    \captionsetup{justification=centering}
    \includegraphics[width=0.48\textwidth,height=0.36\textwidth]{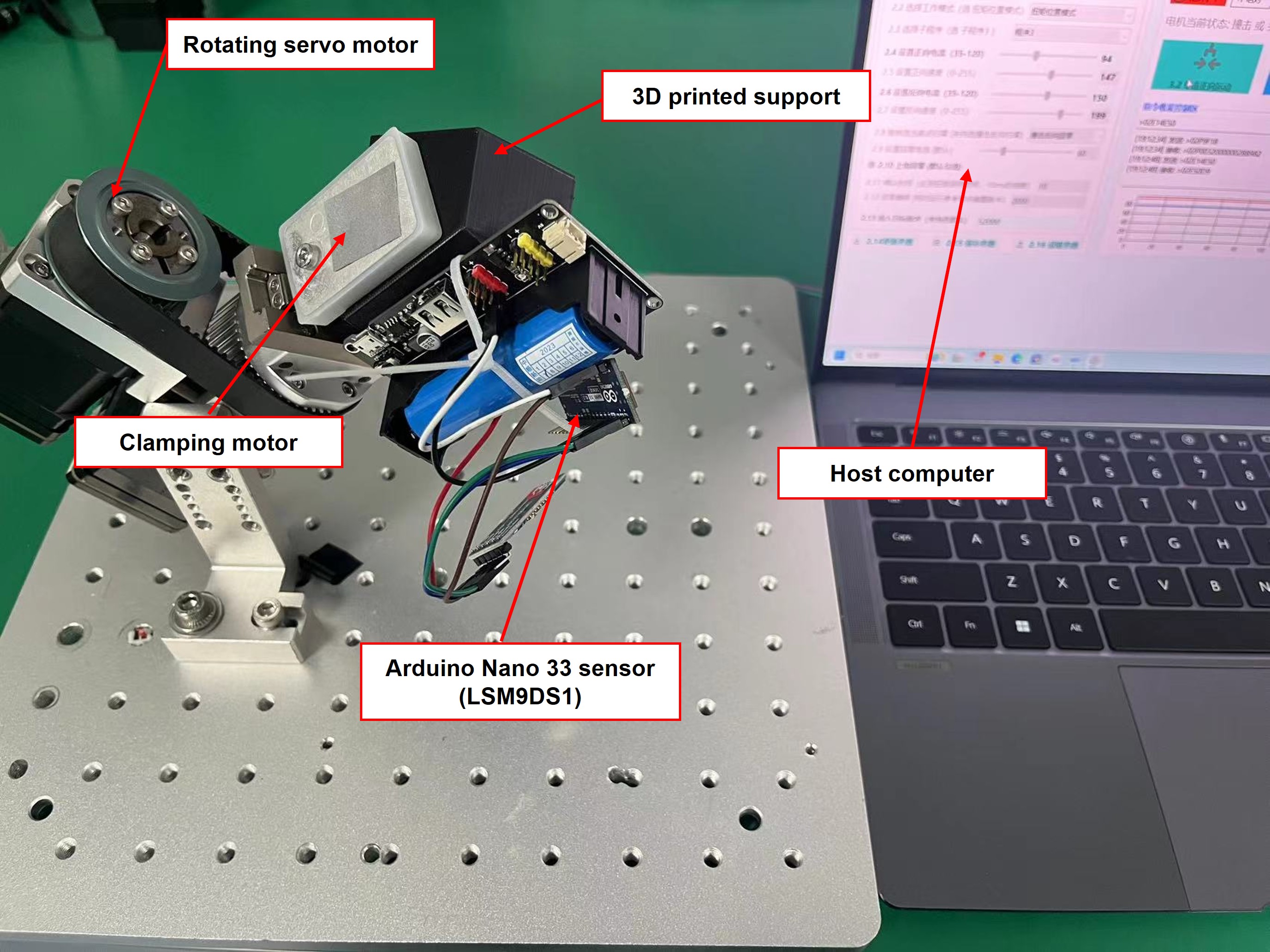}
    \caption{Experimental system for the gyroscope calibration on a servo motor.}
    \label{fig:rotate_table}
\end{figure}

In this section, we undertake the verification of the proposed calibration method via a widely utilized low-cost MEMS micro-IMU, specifically the LSM9DS1 manufactured by STMicroelectronics. The rotational apparatus utilized for this verification, as depicted in Figure \ref{fig:rotate_table}, originates from at the College of  Artificial Intelligence and Big Data in Medical Science, Shandong First Medical University. To ensure comprehensive testing, a 3D-printed rotating support, marked in Figure \ref{fig:rotate_table}, is employed to guarantee that each axis of the gyroscope experiences a rotational component during the experimentation process.

The calibration system and installation arrangements are depicted in Figure \ref{fig:rotate_table}. Initially, the LSM9DS1 Inertial Measurement Unit (IMU) is securely affixed to the rotating support, followed by clamping the 3D printed bracket onto the designated clamping jaws. Subsequently, the motor establishes a communication connection with the control software. Finally, an automation script is utilized to systematically gather gyroscope data for analysis.

\subsection{Experiment Design}
The experiment is structured as follows:

1. \textbf{Accelerometer Data Collection}: The accelerometer will be rotated to a random position and maintained stationary for an extended duration to collect stationary data. In order to better estimate the scale factors, it would be better that the four positions evenly distribute around the testing space.

2. \textbf{Gyroscope Data Collection}: Simultaneously, the gyroscope data will be acquired while the system rotates at a constant speed. All data will be meticulously recorded by the host computer for subsequent analysis.

3. \textbf{Data Processing and Calibration}: Following data acquisition, thorough data processing techniques will be applied, including the extraction of scale factors and biases. These parameters will then be transferred to the computer for further calibration analysis.

4. \textbf{Calibration Range Selection}: The calibration interval for the gyroscope will be determined through a randomized selection process. Specifically, the range from $110^{\circ}$ to $200^{\circ}$ will be designated as the calibration range, with each $10^{\circ}$ increment serving as a calibrated speed point.

\begin{figure}[H]
    \centering
    \captionsetup{justification=centering}
    \includegraphics[width=0.5\textwidth,height=0.38\textwidth]{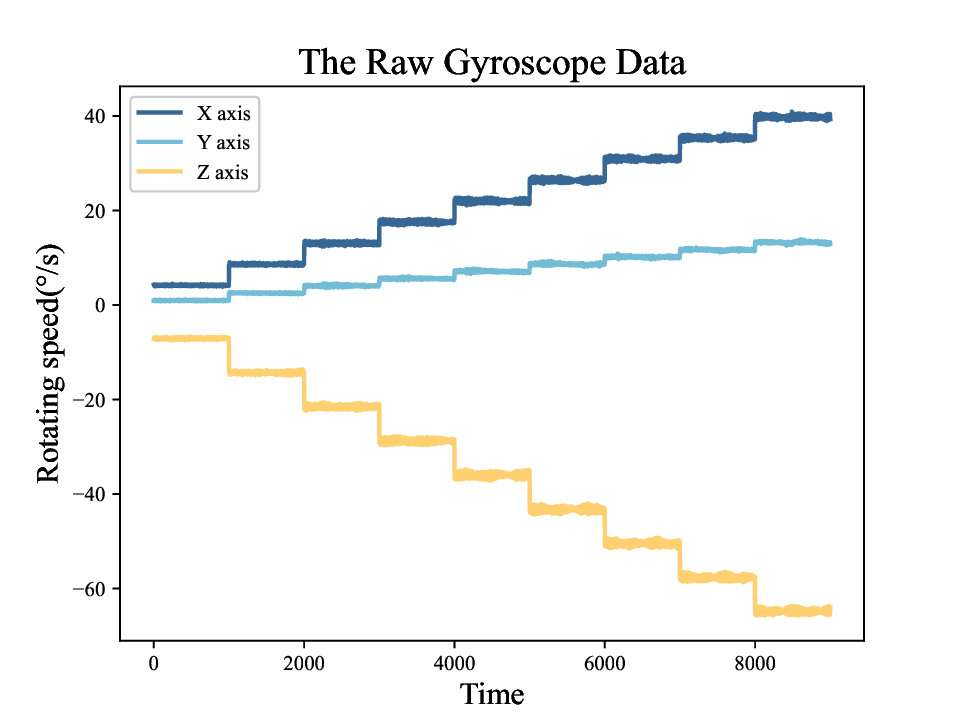}
    \caption{The Raw Data collected from LSM9DS1 gyroscope}
    \label{fig:origin data}
\end{figure}

\begin{figure}[H]
    \centering
    \captionsetup{justification=centering}
    \includegraphics[width=0.5\textwidth,height=0.38\textwidth]{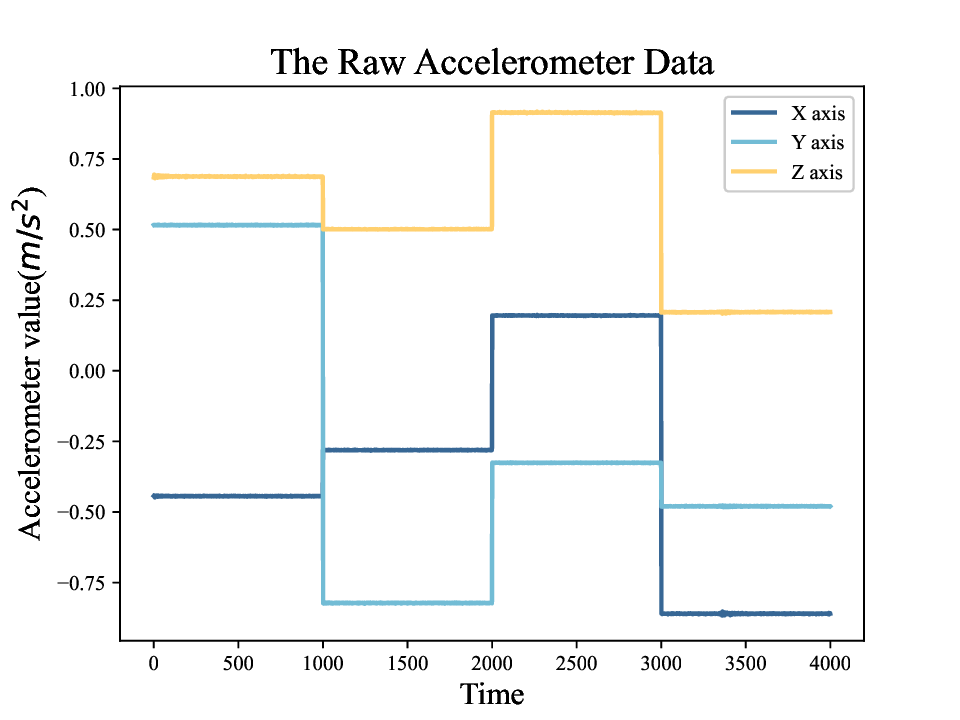}
    \caption{The Raw Data collected from LSM9DS1 accelerometer}
    \label{fig:origin acc data}
\end{figure}

\subsection{Results of Single Speed Point}
\begin{figure}
    \centering
    \captionsetup{justification=centering}
    \includegraphics[width=0.5\textwidth,height=0.38\textwidth]{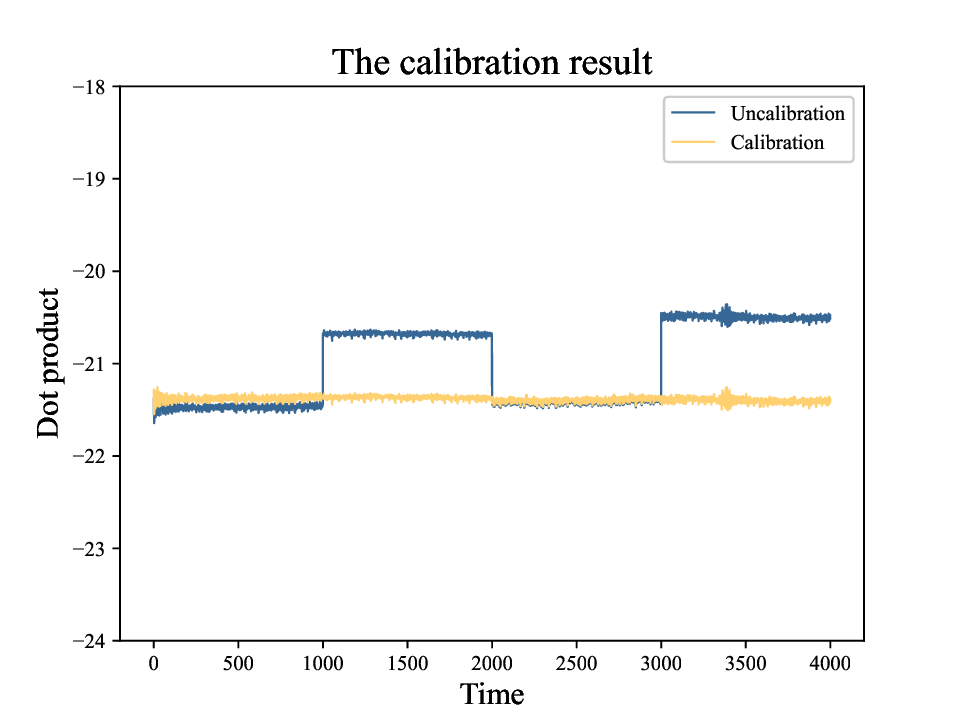}
    \caption{The Calibration result at rotational speed $30^{\circ}/s$}
    \label{fig:singledotproduct}
\end{figure}

Utilizing the calibration method outlined above, a single speed from the range of rotational speeds was randomly selected for experimental calibration analysis. Specifically, a rotational speed of $30^{\circ}/s$ was chosen for the calibration experiment. The calibration results are presented in Figure \ref{fig:singledotproduct}. It is evident from the raw gyroscope data that the three axes exhibit varying degrees of drift. However, following calibration using the method proposed in this paper, the dot product between the accelerometer vector and gyroscope vector becomes consistent across axes.

\subsection{Experimental results of the linearity of the scale factor}
\begin{figure}
    \centering
    \captionsetup{justification=centering}
    \includegraphics[width=0.5\textwidth,height=0.35\textwidth]{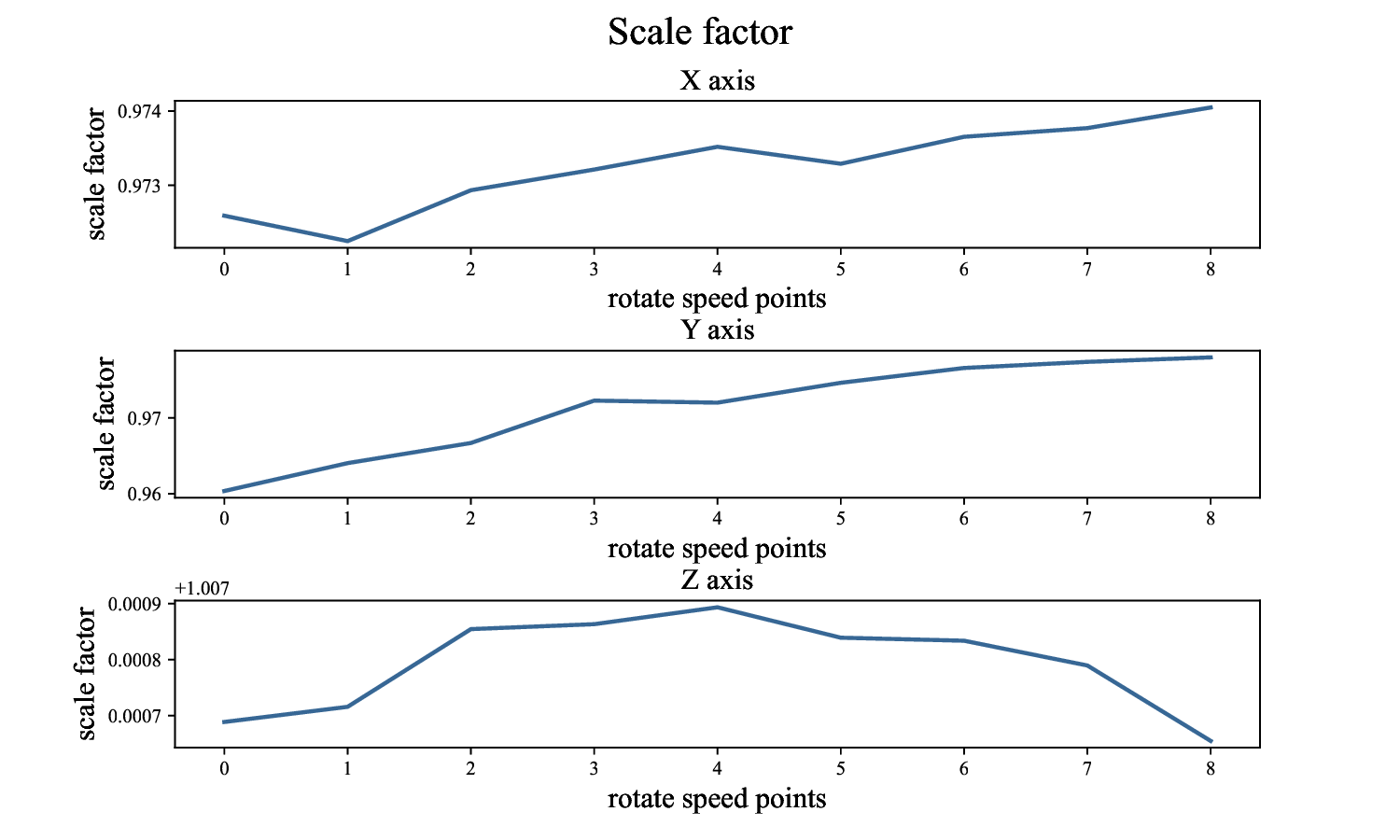}
    \caption{The linearity of gyroscope}
    \label{fig:linearity experiment}
\end{figure}

This subsection is dedicated to presenting the experimental results of determining linearity. The results of linearity obtained from data from the LSM9DS1 gyroscope are illustrated in Figure \ref{fig:linearity experiment}. At each speed point, we applied the recommended calibration method from this paper, obtained the data to be calibrated at each point, and subsequently determined the scale factor for each speed.    

\section{Discussion}

This study introduced a novel approach for calibrating the scale factor and determining the linearity of MEMS triaxial gyroscopes. Through the utilization of linear least squares estimation and leveraging the constant dot product of static accelerometer and rotating gyroscope vectors, we established a method that significantly enhances calibration efficiency and precision. Our simulation studies and experimental validations, conducted with a commercial MEMS IMU and a servo motor, underscore the method's robustness against noise and its ability to maintain accuracy across a range of operational speeds.

The simulation results demonstrated the method's capability to accurately estimate gyroscope parameters, even in the presence of significant noise. Figures \ref{fig:process} and \ref{fig:statistic_result} highlighted the consistency achieved post-calibration, showing a marked improvement in the alignment of the dot product between the accelerometer and gyroscope vectors. This improvement is critical for applications requiring high precision in attitude estimation, where even minor deviations can lead to significant errors in the final output.

Moreover, the experimental results, as depicted in Figure \ref{fig:singledotproduct}, validated the simulation outcomes, further establishing the efficacy of our calibration approach. The experiments were meticulously designed to cover a broad range of rotational speeds, thereby ensuring that the calibration method's effectiveness is not confined to a narrow operational spectrum. Notably, the calibration results at a single speed point, shown in Figure \ref{fig:singledotproduct}, illustrated the method's precision in aligning the scale factors across different axes, thereby significantly reducing the impact of drift.

An important aspect of our method is its capacity to evaluate the linearity of the gyroscope across a specified speed range, as detailed in the experimental results (Figure \ref{fig:linearity experiment}). This capability is paramount for advanced applications where gyroscopes operate under varying conditions, necessitating a calibration method that can adapt to changes in speed without compromising accuracy.

\section{Conclusion}
In this study, we have presented a methodology for calibrating the scale factor and assessing the linearity of gyroscopes. Our approach takes advantage of the relationship between the static accelerometer vector and the rotating gyroscope vector when the axis of rotation remains constant. To streamline the computational demand, we employed a rapid linear least squares estimation technique. Through a comprehensive set of simulations and empirical tests, we have evaluated the efficacy and practicality of our proposed strategy.

The results from our simulations, as illustrated in Figure \ref{fig:statistic_result}, demonstrate the high accuracy with which gyroscope parameters can be estimated. Figures \ref{fig:3d} and \ref{fig:var} further elucidate the impact of noise on the calibration process. To improve calibration precision, we incorporated a method detailed in \cite{7946167} for calibrating accelerometer data, which contributed to the enhancement of the overall calibration outcomes. Our experimental findings affirm the method's efficiency, accuracy, and computational affordability. This technique facilitates swift calibration of each gyroscope axis without necessitating secondary installation procedures. It allows for the initial orientation of the gyroscope to be arbitrary, provided that rotational motion is present on each axis. 

Critically, our method does more than just accurately calibrate each axis of the gyroscope; it also enables the determination of gyroscope parameters across various speed points, thereby allowing for an assessment of gyroscope linearity within any given speed range. This advancement holds significant implications for the precision and applicability of gyroscopes in a wide array of technological and scientific applications.

Finally, the proposed calibration method not only ensures high accuracy in scale factor estimation and linearity determination but also offers a straightforward and computationally efficient approach suitable for on-site applications. The implications of this study are significant, providing a foundation for future research and development in the calibration of MEMS gyroscopes and potentially other types of inertial sensors.





 \bibliographystyle{IEEEtran}
 \bibliography{reference}

\newpage

%
%
%
%

\vfill

\end{document}